# 面向工地风险隐患识别的大模型检索增强框架

李嘉威[1]，杨成业[1]，张尧臣[2]，孙玮琳[1]，孟雷[1*]，孟祥旭[1]

1.山东大学,山东济南 250101； 2.浪潮软件科技有限公司,山东济南 250098

**摘　要**：**目的** 工地风险隐患识别旨在通过自动化技术提升施工现场安全管理水平。现有基于大语言模型的研究分为两类：一是利用图文匹配进行协同推理，但对复杂隐患特征捕捉不足；二是通过专业数据集进行指令微调或多轮对话引导，但存在训练成本高、泛化能力差的问题。**方法** 本文提出一种基于相似案例检索增强的隐患识别方法，通过提示微调技术动态融合外部知识库与检索案例上下文，设计有效的图片检索策略，解决了大模型因领域知识缺失与特征关联弱化导致的误判问题。该方法包括检索库、图片相似度检索和大模型检索增强三个模块，实现了无训练优化下的高效识别。**结果** 实验基于真实施工数据，在多种大模型上测试该方法并进行系统性评估，其中 GLM-4V 的识别正确率提升至 50%，较基线方法显著提高 35.49%，且在多数隐患类别上均表现出性能增益；进一步通过消融实验验证关键模块的有效性，引入学习感知图像块相似度(learned perceptual image patch similarity, LPIPS)算法与图像检索模块中采用的对比语言-图像预训练(contrastive language–image pretraining, CLIP)算法进行对比，结果表明所构建的图像计算检索策略具备优越性。**结论** 本文提出的基于相似案例检索增强方法显著提升了大模型在工地风险隐患识别中的准确率与上下文理解能力,在多类别隐患场景下均表现出良好的泛化性能，为施工现场安全风险智能检测提供了新的理论支撑与技术路径。

# Large model retrieval enhancement framework for construction site risk identification

Li Jiawei[1], Yang Chengye[1], Zhang Yaochen[2], Sun Weilin[1], Meng Lei[1],
Meng Xiangxu[1]

*1.Shandong University,Jinan Shandong 250101，China;2. Inspur Software Technology Co., Ltd., Jinan Shandong* 250098，*China*

**Abstract: Objective** The primary objective of construction site hazard identification is to substantially elevate the safety management standards within operational construction environments by leveraging advanced automation technologies. In recent years, the pervasive adoption of large language models (LLMs) has opened new avenues for research in this critical domain. The overarching goal is to mitigate human error, enhance detection efficiency, and proactively prevent accidents through intelligent systems capable of interpreting complex visual and textual data from construction sites. A meticulous analysis of the current research landscape, based on large language models, reveals that existing methodologies can be broadly classified into two distinct categories, each presenting its own set of advantages and limitations. The first approach leverages the capability of image-text matching to perform collaborative reasoning by integrating visual input with textual hazard descriptions. The second method involves constructing domain-specific datasets to fine-tune large models through instruction tuning or guide them via multi-turn dialogues. The former enhances the alignment between images and semantic representations through multimodal fusion, yet exhibits limitations in capturing complex hazard characteristics. The latter strengthens the model’s analytical depth with domain

*通讯作者

收稿日期:　　　　；　修回日期:

基金项目:基金项目的规范中文全称(项目编号：……) (不同基金之间用分号隔开)

Supported by:基金项目的英文全称（主要基金项目的中英文名称可在学报网站下载中心查找核对）

knowledge infusion but suffers from high training costs and poor generalizability. **Method** To address these limitations, this paper proposes a Risk-Detection Retrieval-Augmented Generation(RDRAG) hazard identification method that dynamically integrates external knowledge bases with retrieved case contexts through prompt tuning, thereby resolving misjudgments caused by LLMs' lack of domain knowledge and weakened feature associations. The proposed architecture is systematically structured into three cohesive and interdependent core modules, each serving a distinct and vital function: The first module is the Retrieval Database Module. This serves as the external knowledge repository, populated with a comprehensive collection of historical construction hazard cases. Each entry within this database is a rich, multi-modal data object comprising visual data (images) and its corresponding textual annotation, which includes a detailed description of the hazard type, its location and contextual information. The integrity, diversity, and relevance of this database are paramount, as it forms the foundational knowledge source for the entire system. The second module is the Image Similarity Retrieval Module. This component is responsible for the efficient and accurate retrieval of the most relevant cases from the database given a new query image from a construction site. At the heart of this module is a powerful vision-language model, specifically the CLIP model. CLIP excels at mapping both images and text into a shared, high-dimensional semantic embedding space. When a new query image is processed, it is encoded into an embedding vector. This vector is then compared against the precomputed embeddings of all images in the retrieval database using a similarity metric. The top-K most semantically similar cases are retrieved, ensuring that the subsequent reasoning steps are informed by visually and contextually analogous examples. The third module is the LLM Retrieval-Augmentation Reasoning Module. This is the central reasoning engine. The retrieved similar cases (both their images and text) are formatted into a structured prompt, providing the large language model (e.g., GLM-4V) with a critical few-shot learning context. This prompt, which also includes the query image, guides the LLM to perform in-context learning. Crucially, this entire framework operates in a training-free manner for the LLM itself, requiring no additional fine-tuning, which guarantees efficiency, reduces computational overhead, and enhances scalability and ease of deployment. **Results** A rigorous empirical evaluation was conducted to validate the efficacy of the proposed framework. Experiments were systematically performed using authentic, real-world construction site data, encompassing a wide variety of hazard scenarios and environmental conditions. The framework was subjected to comprehensive testing and systematic assessment across multiple state-of-the-art large language to ensure the robustness and general applicability of the approach. The results were highly promising and demonstrated a substantial quantitative improvement. Most notably, when integrated with the GLM-4V model, the retrieval-augmentation framework achieved a recognition accuracy of 50%. This represents a significant and remarkable improvement of 35.49% over the baseline performance of the vanilla GLM-4V model without retrieval augmentation. Beyond this aggregate metric, a detailed category-wise analysis revealed consistent performance gains across a majority of individual hazard types. This indicates that the method enhances the model's capability universally rather than being biased towards a specific hazard category. Furthermore, ablation studies were designed, introducing the LPIPS algorithm to compare with the CLIP algorithm used in the image similarity retrieval module. The results demonstrated the clear superiority of the CLIP-based semantic retrieval strategy over the LPIPS-based perceptual strategy in this specific task. **Conclusion** In conclusion, the retrieval-augmented method based on similar cases, as proposed in this paper, delivers a significant breakthrough in automated construction site safety monitoring. It tangibly and markedly enhances the key performance indicators of large language models—namely, accuracy and contextual understanding ability—in the complex task of hazard identification. The framework demonstrates robust generalization performance across a wide spectrum of multi-category hazard scenarios, effectively addressing the core limitations of previous approaches related to training cost and adaptability. The training-free nature of the approach makes it particularly attractive for real-world deployment, offering a scalable and sustainable solution for enhancing on-site safety protocols.

## 0 引 言

工地风险隐患识别（Ballal S 等，2024）旨在通过自动化手段替代传统人工巡检，提升施工安全。当前主流方法基于计算机视觉技术，如目标检测与图像分类（Soumya A 等，2024），但存在泛化能力不足、识别精度低等问题。多模态大模型的发展为隐患识别提供了新思路，然而，由于缺乏工地场景专业知识，模型在复杂环境中易出现误判(Guo Yuanfang 等，2025)。因此，增强模型对隐患场景的理解能力成为核心挑战。

现有研究主要分为两类：一是利用图文匹配（Wang J 等，2024）能力，结合图像与隐患描述进行协同推理；二是通过构建专业数据集（Yu X 等，2024），对大模型进行指令微调或多轮对话引导。前者通过多模态对齐提升图像与语义匹配，但对复杂隐患特征把握有限；后者通过领域知识增强模型分析深度，但存在训练成本高、通用性差的问题。因此，现有方法在领域知识适配性与上下文关联性方面仍存在亟待突破的技术瓶颈。

为在避免高成本微调的同时增强大模型对复杂隐患识别的准确性与领域适应性，引入外部知识源并实施检索增强已成为一种值得探索的解决路径。该方法能够有效结合已有知识库与实时检索机制，在不显著增加训练负担的前提下有效提升模型的上下文感知能力。

本文提出一种基于相似案例检索增强的风险隐患检测识别方法，通过提示微调（Kim T T 等，2025）技术动态融合外部知识库（Zhu L 等，2024）与检索案例上下文，设计有效的图片检索策略，缓解现有多模态大模型因领域知识缺失与特征关联弱化导致的误判问题，如图 1。方法包括三个核心模块：1）检索库模块，构建结构化隐患案例数据库；2）图片相似度检索模块，基于 CLIP（Ghosh A 等，2024）模型定位最相关案例；3）大模型检索增强模块，通过提示微调生成准确隐患类别与描述。

为了验证模型的有效性，本文基于真实施工工地采集数据构建了测试集， 选取 GLM-4V（GLM T 等，2024）、ChatGPT-4o（Lewandowski M 等，2024）以及 DeepSeek-VL2（Wu Z 等，2024）三种主流多模态大模型进行对比评估。实验从识别准确率、误判率、上下文理解能力等方面展开验证。实验结果表明，该方法在多种场景下均显著提升了模型对复杂隐患的识别能力，尤其在难以直接判断的隐患图像中表现出更高的理解深度与判别稳定性。具体来说，本文的主要贡献如下：

1）本文提出了一种基于相似案例检索增强的风险隐患识别框架，创新性地融合大模型提示学习与实例检索机制，为提升大模型在隐患识别任务中的准确性提供了新路径；

2）在算法设计上构建了即插即用的检索增强模块，通过提示微调策略实现大模型的无训练优化，并使其能够在无需额外训练的前提下，快速适应风险隐患识别任务；

3）实验部分我们系统评估了不同大模型在实际场景的识别表现，明确了检索增强在提升模型泛化能力与解释能力方面的优势，为后续多模态大模型工业安全领域的应用提供了理论支持和实践参考。

## 1 相关工作

### 1.1 传统隐患识别方法

工地风险隐患识别任务旨在通过对施工现场潜在危险与隐患的排查、识别与评估，及时采取防范措施，确保工地安全并预防事故发生。传统方法主要依赖安全管理人员的经验判断，通过定期巡查发现隐患并进行整改，然而该方法存在人工疏漏、重复性高以及无法实时监控等局限性。

近年来，基于物联网技术（Wang X 等，2022）与计算机视觉技术（Hou X 等，2023）的隐患识别方法成为研究热点。基于物联网技术的方法通过部署传感器网络，实时监测环境、设备状态及人员行为等数据，结合数据分析技术进行隐患识别。该方法能够动态捕捉施工现场信息，迅速发现潜在风险，但存在成本高、设备需求量大及对数据管理与处理能力要求较高等问题。基于计算机视觉技术的方法则依托图像或视频数据，利用图像处理、目标检测、语义分割及行为识别等技术自动识别安全隐患。其核心在于特征提取，包括传统方法（如尺度不变特征转换(scale invariant feature transform, SIFT)（Arooj S 等，2024）、方向梯度直方图（histogram of oriented gradient, HOG）（Zhang L 等，2024））与深度学习方法（如卷积神经网络（convolutional neural networks, CNN）（Hou Zhiqiang 等，2025））。尽管该方法提升了隐患识别的自动化程度，但其性能受限于图像质量与算法精度，且对计算资源需求较高。

### 1.2 基于大语言模型的隐患识别方法

近年来，大语言模型（Large Language Models, LLMs）在工业隐患识别中的应用逐渐受到关注。LLMs 凭借其强大的语义理解与上下文推理能力（Cui Y 等，2024），能够从文本、图像、传感器数据等多模态信息中提取关键特征，实现对潜在隐患的精准识别。然而，数据稀缺、领域适应性不足以及多模态融合复杂性等问题仍制约其广泛应用。现有研究通过少样本学习（Zeng W 等，2024）、多模态对齐（Wang F 等，2024）

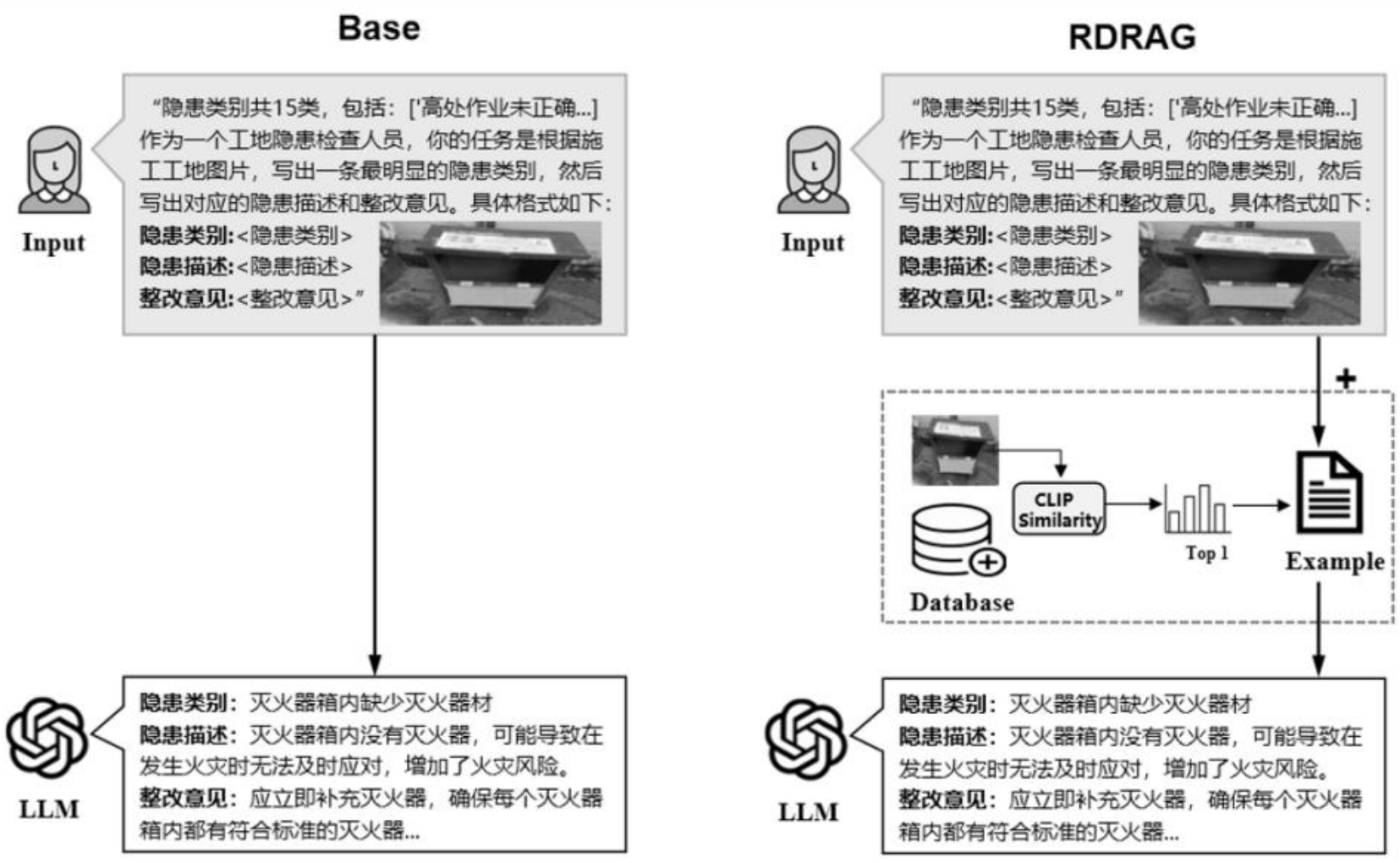

图 1 基于相似案例的检索增强方法

Fig 1 Retrieval-Augmented Method Based on Similar Cases

等技术部分缓解了这些挑战，但仍需要进一步进行优化以提升模型性能。

多模态对齐通过将视觉、文本、传感器等多源数据映射到统一语义空间，实现跨模态特征的协同优化，提升风险识别的准确性与鲁棒性。常用方法包括对比学习（Wei Y 等，2024）、跨模态注意力机制（Luo Y 等，2024）和语义增强表示。例如，Myriad（Li Y 等，2024）结合视觉专家模型与对比学习优化工业场景中的风险识别，大语言模型增强 CLIP 框架（Large Language Model to CLIP ,LLM2CLIP）(Huang W 等，2024）通过语义增强表示与对比学习融合视觉和语言特征。尽管多模态对齐提升了性能，但其计算复杂度较高，对硬件资源需求较大。少样本学习利用预训练模型的泛化能力，在有限标注数据下快速适应新任务，减少对大规模标注数据的依赖。常见方法包括元学习、提示学习与迁移学习。AnomalyGPT（Gu Z 等，2024）与 FedITD（Wang Z Q 等，2024）通过提示学习与迁移学习，在少样本条件下实现精准的工业异常检测。然而，少样本学习可能导致模型在新场景中的泛化能力下降，需针对性调整以提高性能。

如图 1，本文提出了一种基于相似案例检索的增强大语言模型性能的方法，通过构建检索库并匹配与目标相似度最高的案例作为参考，在计算资源消耗较低的情况下实现了隐患检测任务中的少样本学习。该方法的核心在于利用检索机制从小规模案例库中筛选出与当前任务最相关的样本，作为上下文信息输入模型，从而提升模型在低资源场景下的泛化能力和任务适应性。通过引入案例检索增强策略，模型能够有效利用外部知识库中的结构化信息，减少对大规模标注数据的依赖，同时显著降低计算复杂度，为资源受限环境下的高效部署提供了可行方案。

## 2 问题描述

在一个多模态隐患识别系统中，数据集 $\boldsymbol{D}=(\boldsymbol{I_1},C_1,L_1),(\boldsymbol{I_2},C_2,L_2),\cdots,(\boldsymbol{I_N},C_N,L_N)$ 包含了 $N$ 对多模态数据，其中 $\boldsymbol{I_i}$ 表示工地施工图片，$C_i$ 表示隐患描述的文本信息，$L_i$ 表示该隐患类别。由于数据集中包含了图像和文本信息，我们使用 $\boldsymbol{I_i^S}$ 和 $C_i^S$ 来分别表示图像和隐患描述的文本部分。

与传统的需要大规模训练数据的多模态模型不同，我们提出了基于检索增强生成的无训练优化框架，依赖于从数据集中检索出与当前图像 $\boldsymbol{I_i}$ 最相似的历史案例，然后将这些检索到的案例的隐患描述与当前图像共同作为 prompt 输入给大语言模型，生成最终的隐患类别 $L_i'$ 和描述 $C_i'$ 。这个过程不涉及任何训练步骤，因此大语言模型在此过程中并不需要进行参数优化。

具体而言，对于每个输入样本 $(\boldsymbol{I_i},C_i,L_i)$，先从数据集中检索出 $K$ 个与当前图像 $\boldsymbol{I_i}$ 最相似的历史样本 $\{(\boldsymbol{I_j},C_j,L_j)\mid j\in 1,\ldots,K\}$，然后将这些检索到的隐患描述片段 $\{C_j\mid j\in 1,\ldots,K\}$ 与当前图像 $\boldsymbol{I_i}$ 一同输入传递给大语言模型。生成模型依据输入生成当前图像的隐患类别 $L_i'$ 和隐患描述 $C_i'$ 。

用 $L_i'$ 和 $C_i'$ 与实际隐患类别 $L_i$ 和描述 $C_i$ 之间的差异来评估结果生成质量。具体来说，模型的目标是通过检索得到最相似的案例，并通过这些相似案例来辅助生成隐患类别和描述，从而提高识别的

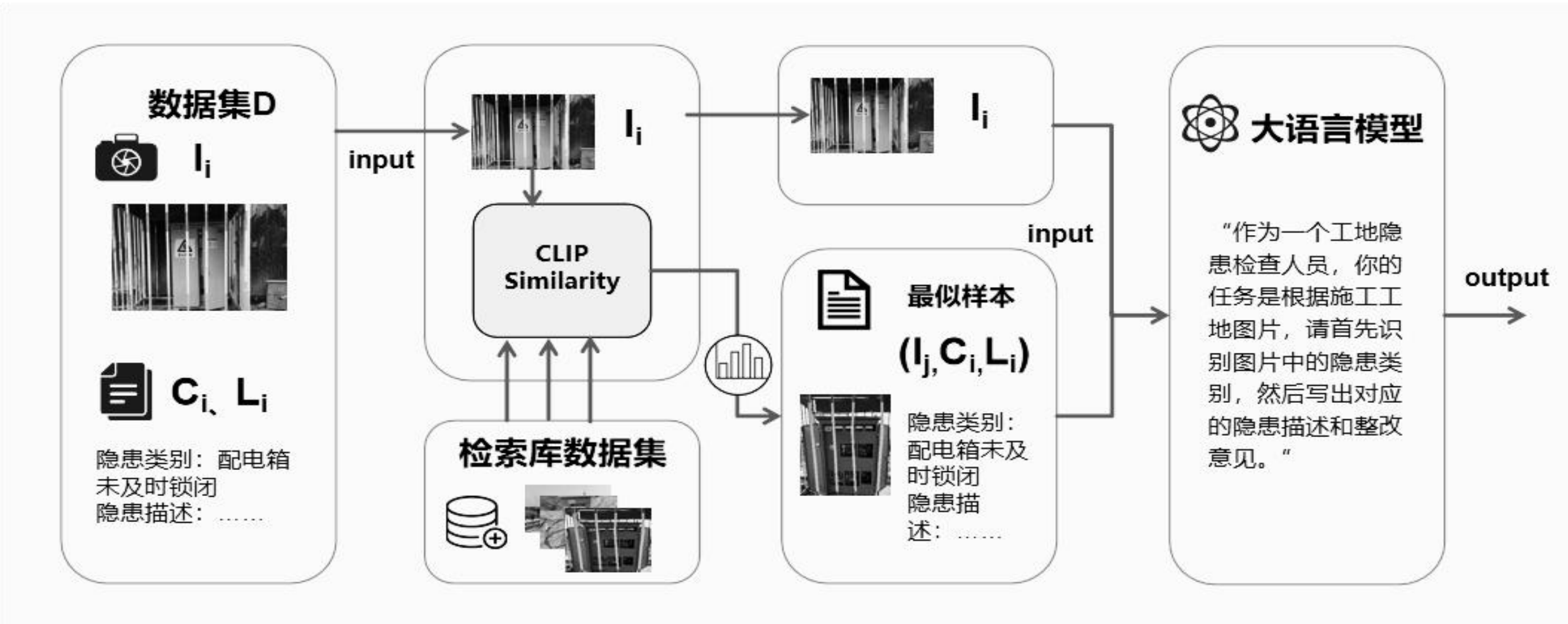

图 2 基于检索增强生成的无训练优化框架

Fig.2 Training-Free Optimization Framework for Retrieval-Augmented Generation

准确性和实用性。可以表示为

$$L_i', C_i' = f(\boldsymbol{I_i}, A\tilde{C}_j \mid j \in 1, \ldots, K) \qquad (1)$$

其中 $\tilde{C}_j$ 表示从相似样本中检索到的隐患描述片段。最终，生成的隐患类别 $L_i'$ 和描述将 $C_i'$ 与实际类别 $L_i$ 和描述 $C_i$ 进行比较，从而评估该检索增强生成过程的效果，确保生成的结果在每个实例中尽可能接近真实标注。

# 3 方法

我们提出了一种基于检索增强生成的无训练优化框架（risk-detection retrieval-augmented generation, RDRAG），其核心结构如图 2 所示。该框架旨在应对现有多模态方法在隐患识别中存在的领域知识缺失与特征关联弱化问题，同时避免高成本微调与泛化性弱的局限。RDRAG 不依赖额外训练，而是通过引入外部知识库与相似案例检索机制动态构建提示信息，从而增强模型对复杂隐患的判别能力与语义关联性。该方法既保持了模型原有的推理泛化能力，又显著提升了在专业场景中输出的准确性与可靠性。

RDRAG 包含三个主要模块，分别是提示词（Prompt）设计（Cao J 等，2024）、大模型检索增强和相似案例检索。通过以上三个模块紧密结合，RDRAG 框架能够在无需大规模训练的情况下，实现对多模态数据的有效理解与生成，极大提升了隐患识别任务的准确性与可靠性。

## 3.1 提示词（Prompt）设计

在 RDRAG 框架中的 Prompt 设计部分，我们致力于通过精心设计不同的提示格式，控制大模型的输出结果。通过添加不同类型的信息到提示中，我们能够引导模型更准确地生成与工地隐患相关的类别、描述和整改意见。为了实现这一目标，我们提出四种不同的提示设计方式，并通过实验定量数据确立最适合我们任务需求的设计。四种提示设计方式包括：

（1）基础指令（type1）：不添加任何信息，只提出要求。只提供了对模型的基本任务要求：“作为一个工地隐患检查人员，你的任务是根据施工工地图片，写出一条最明显的隐患类别，然后写出对应的隐患描述和整改意见。”

（2）类别引导（type2）：添加类别信息。在这种设计中，我们将隐患类别明确列出，要求模型从中选择最相关的类别：“隐患类别的共 15 类包括：['高处作业未正确使用安全带', '基坑支护措施不到位', '灭火器未按规定要求放置', '配电箱未及时锁闭', '起重吊装设备钢丝绳磨损、断丝严重，搭接长度不足', '汽车吊、随车吊、泵车支腿未全部伸出、未垫枕木进行作业', '设备安全防护设施、装置缺失或失效', '未按规定穿戴反光安全服', '未按规定配置灭火器、消防设施等', '未按规定设置接地线或接地不良','现场防护栏等安全防护设施缺失、破损或设置不规范']。”

（3）格式规范（type3）：添加输出格式信息。这种设计进一步优化了输出结构，提供了明确的格式要求：“具体格式如下：隐患类别:<隐患类别>;隐患描述:<隐患描述>;整改意见:<整改意见>”通过这种格式化设计，我们能确保生成的隐患类别、描述和整改意见按标准格式呈现，提高了输出的可读性和一致性。

（4）复合增强（type4）：同时添加类别和格式信息。 “隐患类别的共 15 类包括：['高处作业未正确使用安全带', '基坑支护措施不到位', '灭火器未按规定要求放置', '配电箱未及时锁闭', '

起重吊装设备钢丝绳磨损、断丝严重，搭接长度不足', '汽车吊、随车吊、泵车支腿未全部伸出、未垫枕木进行作业', '设备安全防护设施、装置缺失或失效', '未按规定穿戴反光安全服', '未按规定配置灭火器、消防设施等', '未按规定设置接地线或接地不良','现场防护栏等安全防护设施缺失、破损或设置不规范']。具体格式如下：隐患类别:<隐患类别>;隐患描述:<隐患描述>;整改意见:<整改意见>”这种设计结合了类别信息和格式要求，既能限制模型的输出类别，又能确保生成结果的格式规范，但同时也需要实验验证过于复杂的信息是否会影响大模型的输出结果。

### 3.2 相似案例检索

相似案例检索模块的核心目标是通过 CLIP 模型从数据集中检索出与当前图像$\mathrm{I_i}$最相似的历史案例。CLIP 是一个将图像和文本映射到共同嵌入空间的模型，通过计算图像和文本之间的相似度来实现跨模态匹配。以下是该模块的详细算法描述。

首先，通过 CLIP 模型对输入图像$\boldsymbol{I}_i$和历史图像$\boldsymbol{I}_j$进行特征提取，CLIP 模型会将每张图像映射到一个嵌入空间$f(\boldsymbol{I}_i)$和$f(\boldsymbol{I}_j)$，并将图像和对应的文本描述映射到同一共享空间, 具体函数如下：

$$f(\boldsymbol{I}_i)=\mathrm{CLIP}(\boldsymbol{I}_i),\quad f(\boldsymbol{I}_j)=\mathrm{CLIP}(\boldsymbol{I}_j) \tag{2}$$

这里，$f(\boldsymbol{I}_i)$和$f(\boldsymbol{I}_j)$表示图像$\boldsymbol{I}_i$和$\boldsymbol{I}_j$在 CLIP 嵌入空间中的向量表示。接下来，通过计算图像特征向量之间的余弦相似度衡量当前图像$\boldsymbol{I}_i$和历史图像$\boldsymbol{I}_j$之间的相似性。

$$\mathrm{Sim}(\boldsymbol{I}_i,\boldsymbol{I}_j)=\frac{f(\boldsymbol{I}_i)\cdot f(\boldsymbol{I}_j)}{\|f(\boldsymbol{I}_i)\|\ \ \|f(\boldsymbol{I}_j)\|} \tag{3}$$

其中，$\mathrm{Sim}(\boldsymbol{I}_i,\boldsymbol{I}_j)$表示当前图像$\boldsymbol{I}_i$与历史图像$\boldsymbol{I}_j$之间的余弦相似度，$f(\boldsymbol{I}_i)$和$f(\boldsymbol{I}_j)$分别是图像的特征向量，$\|\cdot\|$表示向量的 L2 范数。通过计算每个历史图像相似度后，我们对所有历史样本进行排序，选择与当前图像最相似的$K$个历史案例$(\boldsymbol{I}_j,C_j,L_j)|\ j\in\{1,2,\ldots,K\}$作为候选样本。

$$\{(\boldsymbol{I}_j,C_j,L_j)|\ j\in\{1,2,\ldots,K\}\}=\text{Top-K}(\mathrm{Sim}(\boldsymbol{I}_i,\boldsymbol{I}_j)) \tag{4}$$

其中，Top-K 操作从所有历史样本中选择前$K$个相似度最高的样本。最后，返回检索到的$K$个历史案例，包含图像$\boldsymbol{I}_j$、隐患描述$C_j$和隐患类别$L_j$，供下游模块使用。

### 3.3 大模型检索增强算法

大模型检索增强模块的目标是将从相似案例检索模块中获得的$K$个隐患描述$C_j|\ j\in 1,2,\ldots,K$和当前图像$\boldsymbol{I}_i$一同输入到多模态大语言模型中，以生成最终的隐患类别$L_i'$和隐患描述$C_i'$。

首先，将当前图像$\boldsymbol{I}_i$和从相似案例检索到的$K$个隐患描述$C_j|\ j\in 1,2,\ldots,K$组成一个输入提示。该提示结合了图像和与之相关的历史隐患描述，以帮助多模态大语言模型生成准确的输出。

$$\mathrm{Prompt}_i=\mathrm{Concat}(\boldsymbol{I}_i,\{C_j|\ j\in\{1,2,\ldots,K\}\}) \tag{5}$$

其中，Concat 表示将图像和隐患描述拼接成一个完整的输入。接着将提示$\mathrm{Prompt}_i$输入多模态大语言模型（例如 GPT 或其他类似语言模型）中，模型根据输入图像和文本描述生成当前图像的隐患类别$L_i'$和隐患描述$C_i'$。

$$L_i',C_i'=\mathrm{LM}(\mathrm{Prompt}_i) \tag{6}$$

这里，LM 表示多模态大语言模型，$L_i'$和$C_i'$分别表示模型生成的隐患类别和描述。

为了更精确地衡量大模型在隐患识别任务中的表现，我们引入以下三种评估指标对生成结果进行系统性评估。

Category Accuracy 指标用于衡量大模型所预测的隐患类别$L_i'$与真实类别$L_i$之间的一致性，反映了模型在隐患分类任务上的判别能力；BERT Similarity 指标用于衡量模型生成的隐患描述$C_i'$和真实隐患描述$C_i$之间的语义相似性，使用预训练的 BERT 模型对两个文本进行编码并计算余弦相似度，评估生成内容在深层语义层面的贴合程度；而 TF-IDF Similarity 指标用于衡量生成文本$C_i'$和真实描述$C_i$在关键词层面的重合程度，利用 TF-IDF 向量表示两个文本，并计算余弦相似度，评估二者在关键信息覆盖方面的匹配度。以上各项指标的具体计算公式将在实验章节的评估指标部分予以详述。

RDRAG 方法的各个关键模块，旨在结合相似案例检索和大模型检索增强，优化多模态隐患识别系统的准确性和可靠性。在无训练优化框架下，RDRAG 方法不依赖于大规模训练数据，通过动态调整提示（prompt）和借助相似案例来增强大语言模型的表现。

# 4 实验

## 4.1 数据集

为了验证添加相似案例检索对大语言模型在隐患识别任务中性能提升的有效性，我们选取了省高速施工真实图片并以此构建了 Rwecd 数据集进行实验研究。该数据集总共包含 325 张隐患图片样本，涵盖 15 种不同的隐患类别，确保数据覆盖了目标工业场景的主要隐患类别和典型环境，每张图片均标注了其对应的隐患类别及隐患描述，用于与大模型的输出结果进行对比，以评估识别准确性。在实验设计中，我们采用分层抽样策略，从数据集中抽取 105 张图片样本构建案例检索库，确保各类别的样本分布均衡；其余 220 张图片样本作为测试集，用于评估模型在未见数据上的泛化能力。

## 4.2 评估指标

为了更精确地衡量大模型在隐患识别任务中的表现，我们引入以下三种评估指标对生成结果进行评估：

Category Accuracy 该指标用于衡量大模型所预测的隐患类别 $L_i'$ 与真实类别 $L_i$ 之间的一致性，定义如下，其中，$\mathbb{k}(\cdot)$ 是指示函数，当 $L_i' = L_i$ 时取值为$1$，否则为$0$，$N$ 表示总样本数量:

$$\text{CategoryAccuracy} = \frac{1}{N}\sum_{i=1}^{N}\mathbb{k}(L_i' = L_i) \quad (7)$$

BERT Similarity（Wang J 等，2024）该指标用于衡量模型生成的隐患描述 $C_i'$ 与真实隐患描述 $C_i$ 之间的语义相似性，使用预训练的 BERT 模型对两个文本进行编码并计算余弦相似度,定义如下，其中，$f_{\text{BERT}}(\cdot)$ 表示使用 BERT 提取的句向量表示：

$$\text{BERTSim}(C_i', C_i) = \frac{f_{\text{BERT}}(C_i') \cdot f_{\text{BERT}}(C_i)}{\| f_{\text{BERT}}(C_i') \| \; \| f_{\text{BERT}}(C_i) \|} \quad (8)$$

TF-IDF Similarity（Jain S 等，2024）该指标用于衡量生成文本 $C_i'$ 和真实描述 $C_i$ 在关键词层面的重合程度。利用 TF-IDF 向量表示两个文本，并计算余弦相似度,其定义如下,其中 $f_{\text{TFIDF}}(\cdot)$ 表示文本的 TF-IDF 向量表示：

$$\text{TFIDFSim}(C_i', C_i) = \frac{f_{\text{TFIDF}}(C_i') \cdot f_{\text{TFIDF}}(C_i)}{\| f_{\text{TFIDF}}(C_i') \| \; \| f_{\text{TFIDF}}(C_i) \|} \quad (9)$$

最终模型生成效果可以通过以上三个指标综合评估，分别对应类别准确性、语义匹配程度与关键词匹配程度，为无训练优化框架的有效性提供客观度量。

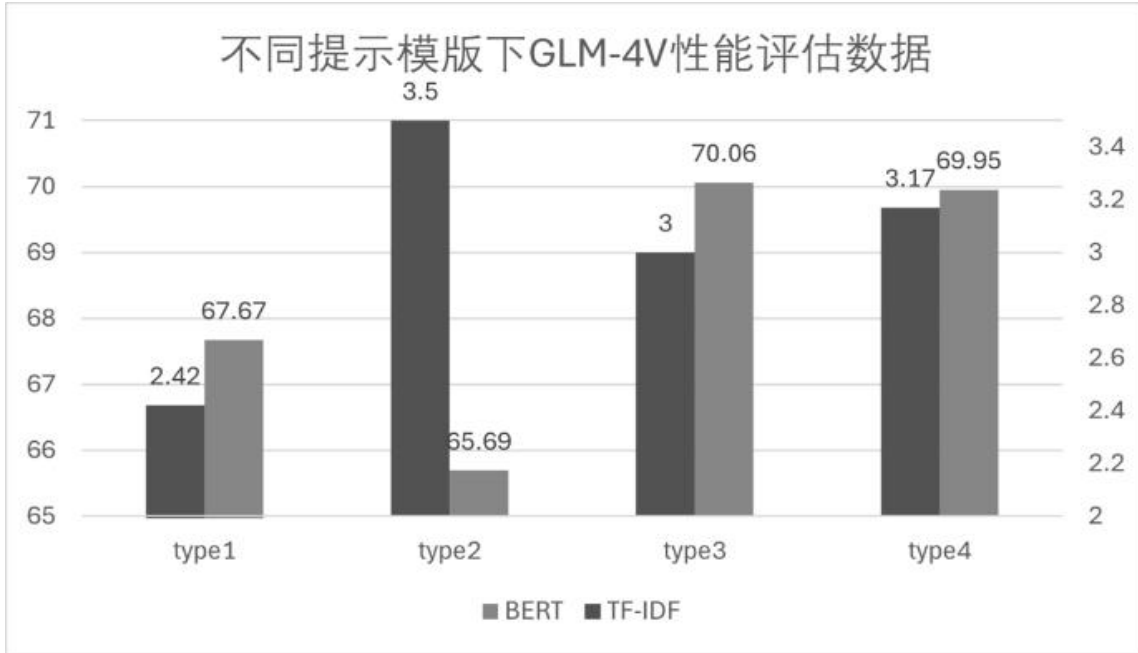

图 3 prompt 对比实验评估数据

Fig 3 Evaluation Data for Prompt Comparison

## 4.3 对比模型

我们在 Rwecd 数据集上展示了 RDRAG 的有效性，并与目前主流的多模态大模型进行对比:

(1)GLM-4V：是通用语言模型智谱 ChatGLM 系列的多模态扩展版本，专注于结合视觉和语言信息进行任务处理;

(2)ChatGPT-4o：是 OpenAI 推出的多模态大语言模型，是 ChatGPT 系列的升级版本，以强大的语言能力为核心，扩展至多模态交互;

(3)Deepseek-vl2：是 DeepSeek 团队开发的多模态大语言模型，专注于视觉与语言的深度融合，适合复杂检索与生成任务.

## 4.4 prompt 对比分析

为系统评估提示模板设计对多模态大模型性能的影响,我们基于 GLM-4V 设计并展开四组对比实验，分别添加四种不同的提示方式（Type1-Type4），在安全隐患图像识别任务上开展对比研究，并通过计算 BERT Similarity 和 TF-IDF Similarity 评估不同提示模版下识别结果的语义匹配程度与关键词匹配程度，并以此选择最优的 prompt 方案。

四组实验组评估数据如图 3 所示，我们得到以下结果并作出分析：

1）Type1 提示模板在两项关键评估指标上均表现欠佳，反映出其生成结果的一致性与判别准确性存在显著不足。该模板采用最简结构，未引入任何约束条件，适用于无需额外上下文引导的任务场景。然而，由于其缺乏明确的语义引导，模型输出呈现较高的随机性，甚至无法稳定判别目标类别。这一现象表明，在复杂视觉理解任务中，过于简化的提示设计可能导致模型性能的显著退化。

2）Type2 模板通过添加类别信息显著提升了关键词匹配度，但语义相似性却出现下降。

这一矛盾性结果表明，类别约束虽能有效缩小模型输出空间，减少类别误判，但由于缺乏结构化输出要求，生成内容的语义连贯性与格式一致性仍存在缺陷。该发现印证了提示需在约束强度与表达自由度间寻求平衡。

**表 1 实验结果数据**

**Table 1 Experimental Results Data**

| 方法 | 模型 | Acc | BERT | TF-IDF |
|---|---|---|---|---|
| Base | GLM-4V | 14.51% | 69.95 | 3.17 |
| | ChatGPT-4O | 53.54% | 71.67 | 5.75 |
| | Deepseek-vl2 | 14.91% | 68.15 | 2.34 |
| COT | GLM-4V | 17.28% | 70.09 | 3.68 |
| | ChatGPT-4O | 55.08% | 71.30 | 4.64 |
| | Deepseek-vl2 | 12.11% | 66.87 | 2.33 |
| RDRAG | GLM-4V | 50.00% | 77.51 | 11.83 |
| | ChatGPT-4O | 59.09% | 73.81 | 6.40 |
| | Deepseek-vl2 | 36.53% | 72.25 | 6.86 |

3）Type3 模板通过规范化输出格式，实现了四组中最优的语义相似性，但其关键词匹配度较 Type2 有所降低。这一差异凸显了结构化提示的双重效应：一方面，固定输出模板可显著提升结果的可读性与逻辑一致性；另一方面，未明确限定类别范围可能导致模型在高层语义推理时出现偏差。

4）Type4 实验组同时添加类别和格式信息，在保持语义相似性与Type3组结果接近的同时，进一步提高了结果的关键词匹配程度。这一设计结合了类别和格式信息，确保了模型输出的类别选择准确且格式规范。通过明确类别范围，我们减少了错误分类的风险，而格式化的输出增强了结果的结构化和一致性。这对于实际的工地隐患检测任务尤为重要，能够提升生成结果的可用性和后续处理的便利性。因此我们最终选择 Type4 作为我们最终实验的 prompt 方案。

## 4.5 性能评估

在 GLM-4V、ChatGPT-4O 和 Deepseek-vl2 三种多模态大语言模型上，我们设计并部署了三组方案进行实验:首先采用设计好的 prompt，不引入任何辅助方法，作为 Base 对照组；其次，作为传统方法的对照组，我们在原有 prompt 设计中添加思维链(COT)（Miao J 等，2024）进行引导，即要求大模型首先在样本图片中定位关键隐患目标物，再基于目标物对隐患类别进行判断，以测试思维链引导对模型推理能力的提升效果；最后采用 RDRAG 框架，从案例库中提取与样本图像相似度最高的案例，并将其作为附加上下文信息输入模型，以评估检索增强方法对模型识别性能的优化作用。

实验结果如表 1 所示，我们发现在引入 COT 方法后，三种大模型的各项数据均未得到明显的提升，Deepseek-vl2 甚至出现了负提升效果，相比之下采用 RDRAG 后，GLM-4V 和 Deepseek-vl2 的识别准确性有较大幅度提升，同时三种大模型的语义匹配程度和关键词匹配程度也均得到较明显的提升。

**表 2 消融实验评估数据**

**Table 2 Ablation Analysis Results**

| 模型 | 方法 | Acc | BERT | TF-IDF |
|---|---|---|---|---|
| GLM-4V | RDRAG | 50.00% | 77.51 | 11.83 |
| | LPIPS | 43.64% | 77.11 | 9.63 |
| | Base | 37.73% | 76.49 | 6.66 |
| ChatGPT-4O | RDRAG | 59.09% | 73.81 | 6.40 |
| | LPIPS | 42.92% | 74.18 | 7.73 |
| | Base | 59.09% | 73.01 | 6.26 |
| Deepseek-vl2 | RDRAG | 36.53% | 72.25 | 6.86 |
| | LPIPS | 27.85% | 70.44 | 4.85 |
| | Base | 24.20% | 70.17 | 3.31 |

实验结果表明，传统思维链（COT）引导方法在提升大模型隐患识别性能方面效果有限，甚至某些情况下会干扰大模型识别。这一现象可能源于 COT 方法在复杂场景中未能有效解决背景干扰问题，导致模型在定位关键隐患目标物时出现偏差，进而影响后续隐患类别的判断。相比之下，RDRAG 框架通过从案例库中检索相似案例作为附加上下文信息，显著提升了模型的识别性能，该结果验证了 RDRAG 框架在增强模型对复杂场景适应能力方面的有效性，表明检索增强方法能更好地辅助模型理解上下文信息，从而提升隐患识别的精度和鲁棒性。

## 4.6 消融实验

为了评估图片相似度检索模块的有效性以及选取 CLIP 作为计算图片相似度算法的合理性，我们设计实施了消融实验。

LPIPS 是一种基于深度学习的衡量图像相似性指标，强调局部感知相似性，关注细节结构，其应用在图像修复等任务中有较好的表现，但不同于 CLIP，LPIPS 无跨模态能力，因此缺乏图像语义内容的理解能力。

实验设计中，我们首先移除了图片相似度检索模块，并采用随机检索策略，即从检索库中随机抽取案例作为附加上下文信息输入模型，以此构建 Base 组；接着我们用 LPIPS 算法替换掉了图片相似度检索模块中的 CLIP 算法，以此计算搜索相似度最高图片，并构建 LPIPS 组。实验结果如表 2 所示。通过对比分析我们得到如下结果：

1）检索库机制对模型性能的影响：三种多模态大语言模型引入 RDRAG 后评估指标有所提升，其中 GLM-4V 和 DeepSeek-VL2 提升效果较为显著，无论是 LPIPS 还是 CLIP 应用在检索模块中都提高了多模态大语言模型的识别能力，说明提示学习通过优化模型的上下文理解能力，显

著增强了隐患识别的准确性与鲁棒性。

表 3 不同类别间准确率对比

**Table 3 Per-class Accuracy Benchmarking**

| 隐患类别 | 数量 | Base | RDRAG |
|---|---|---|---|
| 1.未按规定穿戴反光安全服 | 4 | 0.00 | 33.33 |
| 2:高处作业未正确使用安全带 | 15 | 46.00 | 33.33 |
| 3:配电箱未及时锁闭 | 30 | 26.00 | 60.00 |
| 4:未按规定配置灭火器、消防设施等 | 20 | 0.00 | 50.00 |
| 5:现场防护栏等安全防护设施缺失、破损或设置不规范 | 25 | 32.00 | 23.53 |
| 6:设备安全防护设施、装置缺失或失效 | 25 | 12.00 | 64.71 |
| 7:起重吊装设备钢丝绳磨损、断丝严重，搭接长度不足 | 25 | 0.00 | 58.82 |
| 8:汽车吊、随车吊、泵车支腿未全部伸出、未垫枕木进行作业 | 30 | 0.00 | 70.00 |
| 9:基坑支护措施不到位 | 12 | 66.67 | 12.50 |
| 10:灭火器未按规定要求放置 | 6 | 33.33 | 0.00 |
| 11:未按规定设置接地线或接地不良 | 28 | 0.00 | 31.58 |
| 12:安全警示标志标识缺失或设置不规范 | 20 | 55.00 | 35.71 |
| 13:灭火器压力不足，灭火器、消防设施等未按规定进行检查、维护 | 25 | 0.00 | 23.53 |
| 14:不符合“三级配电两级漏电保护、一机一闸一漏一箱”要求 | 30 | 0.00 | 60.00 |
| 15:电缆外皮破损或敷设不规范 | 30 | 0.00 | 65.00 |

2）CLIP 方法的有效性：相比于应用 CLIP 的 RDRAG 组，应用 LPIPS 的实验组对多模态大语言模型的提高效果较不明显，甚至在 ChatGPT-4O 模型上准确率出现了负提升的效果，说明 LPIPS 虽然在感知相似性任务中具有较好的表现，但由于缺少跨模态的识别能力，在理解图片内容相似性上，CLIP 是更加有效的选择。

### 4.7 深入分析

为探究本文提出 RDRAG 方法对多模态大语言模型在隐患识别任务中性能提升的有效性，我们以 GLM-4V 模型为例，分别对数据集中 15 个隐患类别进行了系统评估。我们统计了模型在引入 RDRAG 前后不同类别 Category Accuracy 的数据，通过定性分析探究影响 RDRAG 方法在少样本类别中表现的因素，实验数据如表 3 所示。通过深入分析不同类别的准确率变化，我们得到以下结果：

(1)RDRAG 方法的优化效果：在引入 RDRAG 方法后，GLM-4V 模型在大部分隐患类别中的评估指标均呈现显著性提升，尤其是在数据样本较多（如第 3 类、第 8 类）或关键目标物相似（如第 4 类、第 13 类）的类别识别上有着正向提升效果。

(2)少样本类别的表现：对于一些样本数量极少的类别（如第 1 类、第 10 类），RDRAG 方法的优化效果并不稳定，这是由于模型本身难以学习到判别性特征，此时性能上限受到数据制约。

(3)小目标感知问题：大模型在图像细节及上下文环境捕捉方面表现优异，但易受复杂背景干扰，导致在复杂场景中难以精准识别隐患点，而 RDRAG 方法在场景较为复杂的类别（如第 6 类，第 8 类，第 15 类）识别上，展现出了较为显著的优化效果。

## 5 结 论

本文提出了一种基于检索增强生成的无训练优化框架 RDRAG。该框架通过结合提示微调和相似案例检索，解决多模态领域中因特征关联弱化而导致的误判问题。实验表明，所提出的 RDRAG 可以通过检索增强优化大模型的输出，显著提升大模型对隐患的上下文理解能力，从而增强其识别与泛化的检测性能。未来我们将尝试采用更复杂的 RAG（Zhu L 等，2024）提示增强技术提升模型的推理能力，弥补复杂场景下隐患点捕捉能力的不足，对于应对小目标感知的隐患识别任务能力进一步加强。

**参考文献**(References)